\documentclass[nohyperref]{article}

\usepackage{microtype}
\usepackage{graphicx}
\usepackage{subcaption}
\usepackage{booktabs} 
\usepackage{enumitem}

\usepackage{array}
\usepackage{booktabs}
\usepackage{multirow}

\usepackage{hyperref}

\usepackage[accepted]{icml2022}

\usepackage{amsmath}
\usepackage{amssymb}
\usepackage{mathtools}
\usepackage{amsthm}

\usepackage[capitalize,noabbrev]{cleveref}

\newcommand{\ie}{\emph{i.e., }}
\newcommand{\eg}{\emph{e.g., }}

\newcommand{\wrt}{\emph{w.r.t. }}
\newcommand{\cf}{\emph{cf. }}
\newcommand{\aka}{\emph{aka. }}

\newcommand{\Lapl}{\mathbf{\mathop{\mathcal{L}}}}

\newcommand{\Trans}[1]{{#1}^{\top}}

\newcommand{\Mat}[1]{\textbf{#1}}

\newcommand{\Space}[1]{\mathbb{#1}}
\newcommand{\Set}[1]{\mathcal{#1}}

\definecolor{aqua}{rgb}{0.0, 1.0, 1.0}

\theoremstyle{plain}

\theoremstyle{definition}

\theoremstyle{remark}

\usepackage[textsize=tiny]{todonotes}

\icmltitlerunning{Let Invariant Rationale Discovery Inspire Graph Contrastive Learning}

\begin{document}

\twocolumn[
\icmltitle{Let Invariant Rationale Discovery Inspire Graph Contrastive Learning}



\icmlsetsymbol{equal}{*}

\begin{icmlauthorlist}
\icmlauthor{Sihang Li}{ustc-is}
\icmlauthor{Xiang Wang*}{ustc-cs}
\icmlauthor{An Zhang}{nus-next}
\icmlauthor{Ying-Xin Wu}{ustc-ds}
\icmlauthor{Xiangnan He*}{ustc-ds}
\icmlauthor{Tat-Seng Chua}{nus-next}
\end{icmlauthorlist}

\icmlaffiliation{ustc-is}{School of Information Science and Technology, University of Science and Technology of China, Hefei, China}
\icmlaffiliation{ustc-cs}{School of Cyber Science and Technology, University of Science and Technology of China, Hefei, China}
\icmlaffiliation{ustc-ds}{School of Data Science, University of Science and Technology of China, Hefei, China}
\icmlaffiliation{nus-next}{Sea-NExT Joint Lab, National University of Singapore, Singapore}

\icmlcorrespondingauthor{Xiang Wang}{xiangwang1223@gmail.com}
\icmlcorrespondingauthor{Xiangnan He}{xiangnanhe@gmail.com}

\icmlkeywords{Graph Contrastive Learning, Data Augmentation, Invariant Rationale Discovery, machine learning, ICML}

\vskip 0.3in
]



\printAffiliationsAndNotice{} 

\begin{abstract}

    Leading graph contrastive learning (GCL) methods perform graph augmentations in two fashions:
    (1) randomly corrupting the anchor graph, which could cause the loss of semantic information, or (2) using domain knowledge to maintain salient features, which undermines the generalization to other domains.
    Taking an invariance look at GCL, we argue that a high-performing augmentation should preserve the salient semantics of anchor graphs regarding instance-discrimination.
    To this end, we relate GCL with invariant rationale discovery, and propose a new framework, \underline{R}ationale-aware \underline{G}raph \underline{C}ontrastive \underline{L}earning (RGCL).
    Specifically, without supervision signals, RGCL uses a rationale generator to reveal salient features about graph instance-discrimination as the rationale, and then creates rationale-aware views for contrastive learning.
    This rationale-aware pre-training scheme endows the backbone model with the powerful representation ability, further facilitating the fine-tuning on downstream tasks.
    On MNIST-Superpixel and MUTAG datasets, visual inspections on the discovered rationales showcase that the rationale generator successfully captures the salient features (\ie distinguishing semantic nodes in graphs).
    On biochemical molecule and social network benchmark datasets, the state-of-the-art performance of RGCL demonstrates the effectiveness of rationale-aware views for contrastive learning.
    Our codes are available at \url{https://github.com/lsh0520/RGCL}.

\end{abstract}

\section{Introduction}

Self-supervised contrastive learning \cite{he2020momentum,SimCLR} is attracting a surge of interest in graph neural networks (GNNs) \cite{GNNSurvey,Benchmarking}, giving rise to graph contrastive learning (GCL).
It pre-trains a GNN on a large dataset without relying on handcrafted annotations, which promotes the fine-tuning on downstream tasks \cite{DGI,GCC,GraphCL}.
And GCL has a tremendous impact on real-world applications with abundant unlabeled data, such as the discovery of potential drug-drug interactions in drugs' pharmacological activity \cite{wang2021multi}.

Inspecting prior studies on GCL, we can systematize the common paradigm as a combination of two modules: (1) graph augmentation, which creates the augmented views of anchor graphs via various approaches, such as node dropping, edge perturbation, and attribute masking;
and (2) contrastive learning, which maximizes the agreement for two augmentations of the same anchor, while minimizing the agreement for those of two different anchors.
Clearly, graph augmentation is of crucial importance to delineate the discriminative characteristics of graph instances and generate instance-discriminative representations.

\begin{figure}[t]
    \centering
    \includegraphics[width=\columnwidth]{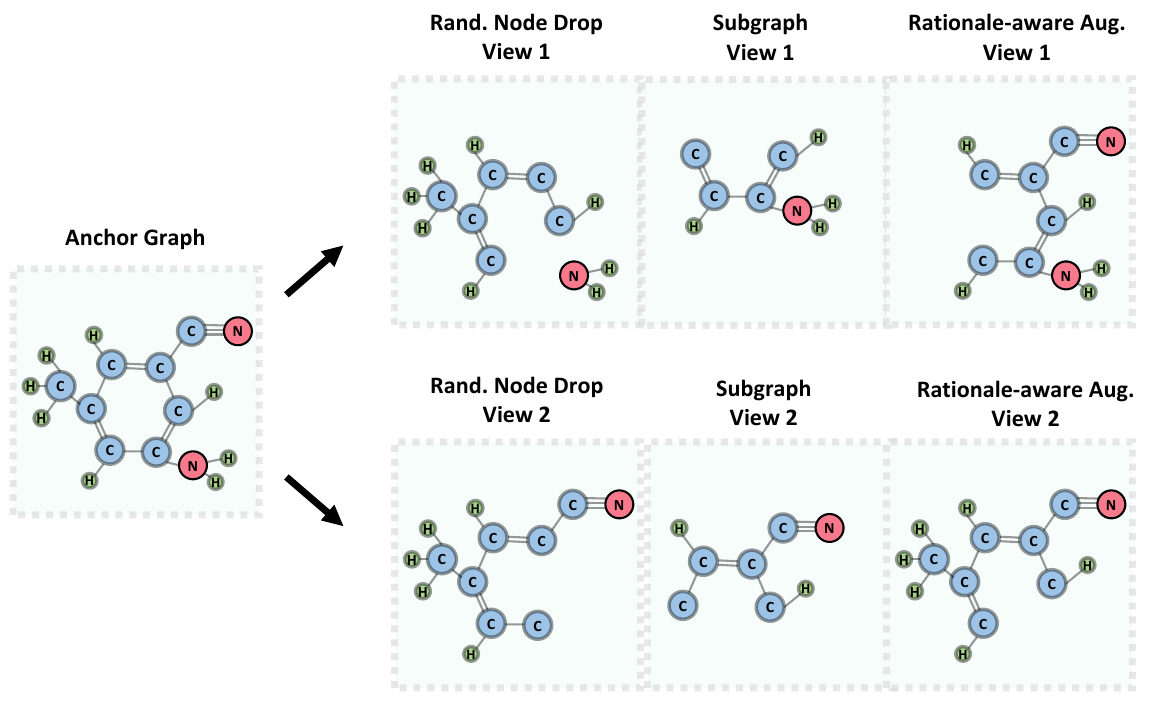}
    \vspace{-20pt}
    \caption{Illustration of graph augmentation in random fashions and rationale-aware fashions. In the molecule graph, the cyano group (\emph{-C$\equiv$N}) works as the rationale to its toxicity.}
    \label{fig:intro_data_aug}
    \vspace{-15pt}
\end{figure}

State-of-the-art GCL methods conduct graph augmentation either in a random fashion \cite{GraphCL, GRACE, GCC} or in a knowledge-guided fashion \cite{motif-GCL, GCA, wang2021multi}.
These two lines suffer from inherent limitations correspondingly:
\begin{itemize}[leftmargin=*]
    \item Graphs usually contain rich saliency information and regularity structure (\eg functional groups in molecule graphs, coalitions in social networks).
    Thus, randomly corrupting their properties (\eg topological structure, node properties, or edge attributes) could lose the discriminative semantics, making the augmented views far from the anchor graphs and misguiding the contrastive learning.
    Figure \ref{fig:intro_data_aug} showcases random augmentations are likely to pervert the cyano group (\emph{-C$\equiv$N}), which is the key feature making the molecule hypertoxic, thus easily derailing the discriminative power of the augmented views.
    

    \item A research line leverages external domain knowledge to identify the salient semantics of graphs and guide the augmentations.
    For example, underscoring the functional groups, such as the cyano group for toxicity in Figure \ref{fig:intro_data_aug}, is beneficial to molecule representation learning \cite{GROVER}.
    Nonetheless, it is prescribed to expensive domain knowledge and easily undermines the generalization performance in unseen domains.
    
\end{itemize}

To address these limitations, we revisit the GCL paradigm from the invariance standpoint \cite{PIRL,ECL}.
Specifically, the contrastive learning module encourages the representation agreement between the anchor graph and its augmented views, \ie the representations are invariant to the augmentations.
Formally, let $g$ be the anchor graph, $A(\cdot)$ be the augmentation function, and $f(\cdot)$ be the encoder network yielding the representations.
We can formulate the goal of invariance as $f(A(g))=f(g)$, where $f(g)$ preserves the salient semantics but contains no information about the way $g$ is augmented.

One naturally raises question: \emph{What augmentations should the representations be invariant to?} To answer it, we relate GCL with invariant rationale discovery (IRD) \cite{DIR,InvRationale,IGV}.
IRD usually involves two modules: (1) rationale discovery, which can be formulated as a function $R(\cdot)$, extracting features guiding the prediction as the rationale; and (2) prediction, which uses the rationale only for supervised prediction.
The prediction is influenced by the rationale, which is invariant regardless of the changes in the rationale's complement.
Considering $f(\cdot)$ as the encoder encapsulating the information to perform prediction, we formulate the aim of IRD as $f(R(g))=f(g)$.
As such, $R(\cdot)$ is supposed to reveal $g$'s critical information, which is substantially consistent with the idea of GCL.
We might assume that the representations should be invariant to the rationale-aware augmentations that preserve instance-discriminative information.

Motivated by the connection of IRD and GCL, we propose a new framework, \underline{R}ationale-aware \underline{G}raph \underline{C}ontrastive \underline{L}earning (RGCL), to automatically discover rationales as graph augmentations.
Specifically, RGCL is a cooperate game between two modules:
(1) rationale generator, which decides fractions to reveal and conceal in the anchor graph, and yields the rationale encapsulating its instance-discriminative information;
and (2) contrastive learner, which makes use of rationale-aware views to perform instance-discrimination of graphs.
The two players aim for the shared goal of achieving semantically-good representations.
On both biochemical molecule and social network benchmark datasets \cite{Molecule-Net, TU-dataset}, extensive experiments demonstrate the promising performance of RGCL to surpass current state-of-the-art GCL methods \cite{GNN-pretrain, GraphCL,graph-l-o-g, AD-GCL}.

\section{Invariance Look at GCL}
We begin with preliminary and related work of GCL and IRD.
We then present the invariance view to relate them.

\subsection{Graph Contrastive Learning (GCL)}
Self-supervised learning has recently achieved promising success in various domains \cite{revisit-ssl-cv,benchmark-ssl-cv,BERT,GPT-3}, thus provoking the interest in GNNs.
Its goal is to pre-train a GNN on a large dataset in a self-supervised fashion, so as to enhance the expressive power of GNN and facilitate its fine-tuning on downstream tasks.
Such pre-training strategies roughly fall into two research lines.
Early studies create augmented views of original graphs and require the learners to predict certain properties from the views, such as graph context \cite{GNN-pretrain}, node attribute \cite{GNN-pretrain} and edge presence \cite{GraphSAGE}.
As such, the resultant representations are covariant to the augmentations \cite{PIRL,ECL}.

Recently, studies on GCL have attracted increasing attention, which is the focus of this work. The most commonly-used criterion for GCL is maximizing the mutual information \cite{infonce} between two augmented views \cite{GCC,DGI,GraphCL}.
It remains within the confines of the two modules: graph augmentation and contrastive learning.

\textbf{Graph augmentation.} 
For an anchor graph instance $g$ in the graph set $\Set{G}$, this module employs the augmentation function $A(\cdot)$ to create $g$'s views as:
\begin{gather}\label{equ:graph-augmentation-paradigm}
    g_{A}=A(g),\quad A\in \Set{A},
\end{gather}
where $A(\cdot)$ is the function sampled from graph augmentation function set $\Set{A}$ to yield the augmented view $g_{A}$. We next elaborate the design of $A(\cdot)$ in prior studies.

Most GCL methods instantiate function $A(\cdot)$ in a random fashion by randomly corrupting the anchor's topological structure, node properties, or edge attributes.
For example, GRACE \cite{GRACE} is a hybrid scheme augmenting on the structure and attribute levels.
GCC \cite{GCC} captures the universal topological properties across multiple graphs.
GraphCL \cite{GraphCL} systematically studies the impact of combining various random augmentations.
Despite promising performance, the intrinsic random nature makes them suffer from potential semantic information loss, thus hardly capturing the salient information.
To tackle this problem, domain knowledge is used to identify salient features in graph augmentations.
Take the biochemistry domain as an example.
MICRO-Graph \cite{motif-GCL} extracts semantically meaningful motifs (\eg functional groups of molecules) to construct informative subgraphs.
More recently, GraphMVP \cite{GraphMVP} utilizes the correspondence between 2D topological structures and 3D geometric views of molecules to perform contrastive learning.
However, heavily relying on domain knowledge severely undermines their generalization to other domains.
In the domain of social networks, GCA \cite{GCA} exploits node centrality to highlight important connected structures, thus recognizing underlying semantic information.
Nevertheless, its intrinsic assumption --- node centrality represents saliency --- could possibly fail in other domains.

\textbf{Contrastive learning.}
Given the augmented views of $g$, contrastive learning sets up an instance-discrimination task, which can be formulated as follows:
\begin{gather}\label{equ:contrastive-learning-paradigm}
    \min_{f} \Lapl_{\text{CL}} = \Space{E}_{g\in\Set{G}}\Space{E}_{A_{1},A_{2}\sim \Set{A}}l_{c}(f(A_{1}(g)), f(A_{2}(g))),
\end{gather}
where $f(\cdot)$ is the GNN encoder to yield representations, $A_{1}$ and $A_{2}$ are two functions sampled from graph augmentation function set $\Set{A}$. $l_{c}(\cdot,\cdot)$ can be instantiated as any contrastive loss function, including NCE \cite{PIRL}, InfoNCE \cite{infonce} and NT-Xent \cite{SimCLR}.

\textbf{Invariance Look.} Optimizing Equation \eqref{equ:contrastive-learning-paradigm} encourages the encoder $f(\cdot)$ to produce the same representation for anchor graph $g$ as for its augmented counterparts.
That is, the representations are invariant to the augmentation \cite{PIRL,ECL}, formally:
\begin{gather}\label{equ:gcl-invariance}
    f(A(g))=f(g),
\end{gather}
which essentially enforces the augmented view $g_A = A(g)$ to preserve the critical information in anchor $g$ that is distinguishable from the other anchors.

\subsection{Invariant Rationale Discovery (IRD)}
Towards interpretability/explainability \cite{GNNExplainer,L2X,RC-Explainer,ReFine} in supervised learning scenarios, IRD \cite{InvRationale,DIR,IGV} finds a small subset of the input features as rationale, which guides and interprets the prediction.
It usually consists of two modules: rationale discovery and classification.

With a slight abuse of notation, we denote the graph, label, rationale and complement variables as the uppercase $G$, $Y$, $R(G)$ and $C(G)$, respectively.
Correspondingly, the lowercase $g$, $y$, $R(g)$ and $C(g)$ represent samples/values of the aforementioned variables.

\textbf{Rationale Discovery.} For a graph instance $g$, this module extracts a substructure of $g$ as $R(g)$, which is the supporting substructure termed rationale and allows a confident classification alone.
Specifically, $R(\cdot)$ is an instantiation of graph augmentation function, preserving critical substructure of $g$.
The rationale can compose salient topological features, node properties, or edge attributes.
For example, DIR \cite{DIR} applies an attention network on $g$'s edges to select the salient edges with top attentive attributions as the rationale of this graph instance.

\begin{figure*}[t]
    \centering
    \includegraphics[width=\textwidth]{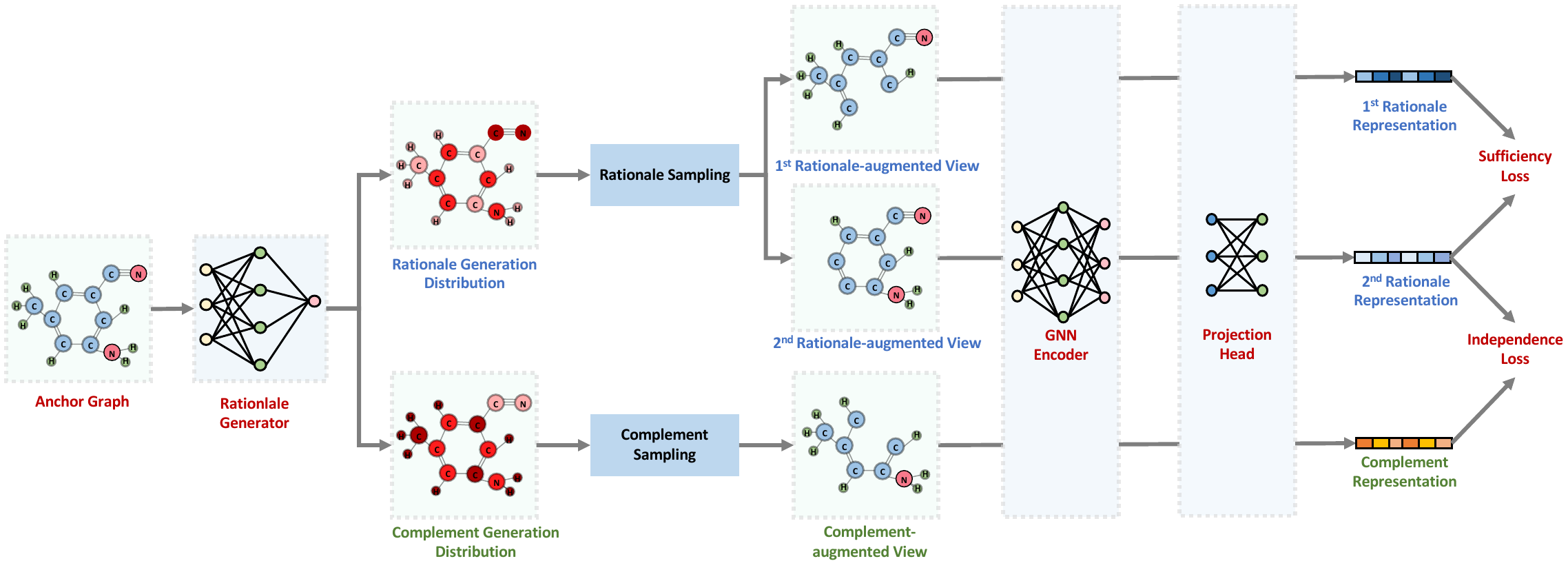}
    \vspace{-10pt}
    \caption{RGCL framework. A rationale generator identifies a discriminative subset of nodes in the original graph. Darker color indicates higher sampling probability in the rationale/complement generation distribution. Two rationales and one complement are generated according to Equation \eqref{equ:rationale-distribution} and Equation \eqref{equ:complement-distribution}. The rationale generator, shared GNN-encoder and shared projector are jointly optimized via minimizing both the sufficiency loss (\cf Equation \eqref{equ:loss-su}) and independence loss (\cf Equation \eqref{equ:loss-in}).}
    \label{fig:framework}
    \vspace{-10pt}
\end{figure*}

\textbf{Classification with Two Principles.}
Owing to the lack of ground-truth rationales, learning to discover rationale is achieved by implementing function $R(\cdot)$ with a network $r(\cdot)$ and approaching the original input $g$'s target label $y$.
In general, the oracle rationale needs to satisfy two principles: sufficiency \cite{DBLP:conf/emnlp/YuCZJ19} and independence \cite{InvRationale,DIR}, which are adopted by the module to push the discovered rationale close to the oracle.
By ``sufficiency'', we mean that the rationale $R(g)$ is sufficient to preserve the critical information of $g$ related to the label $y$, which is formulated as:
\begin{gather}\label{equ:sufficiency-principle}
    p_{Y}(\cdot|R(g))=p_{Y}(\cdot|g),
\end{gather}
where $p_{Y}(\cdot|X=x)$ is conditional probability density/mass function conditioned on the random variables corresponding to input $x$.
By ``independence'', we mean that label variable $Y$ is independent of the complement variable of rationale, $C(G)$, conditioned on the rationale $R(G)$:
\begin{gather}\label{equ:independence-principle}
    Y\perp C(G) | R(G),
\end{gather}
where $\perp$ is probabilistic independence.
These two principles shield the classification from the influence of information outside the rationale, so as to enforce the rationale to collect all discriminative features.
Moreover, they help to prevent the rationale from degeneration \cite{DBLP:conf/emnlp/YuCZJ19}, where trivial information is collected.

To model these principles, IRD usually devises an encoder network $f(\cdot)$ to generate the rationale representations and a subsequent classifier network $\phi(\cdot)$ to predict $g$'s label:
\begin{gather}\label{equ:ird-training-paradigm}
    \min_{\phi,f,r} \Lapl_{\text{IRD}}= \Space{E}_{(g,y)\in \Set{D}}l_{s}( \phi(f(R(g))),y)\\
    \text{s.t.}\quad Y\perp C(G) | R(G),\nonumber
\end{gather}
where $\Set{D}$ is the dataset involving with the pairs of graph instances and target labels; $l_s(\cdot, \cdot)$ measures the supervised loss like cross-entropy.

\textbf{Invariance Look.}
We also take an invariance look at IRD.
First, the sufficiency principle (\cf Equation \eqref{equ:sufficiency-principle}) enforces the encoder $f(\cdot)$ to refine the same information for rationale $R(g)$ as for its original graph $g$.
Second, the independence principle (\cf Equation \eqref{equ:independence-principle}) naturally makes classification insensitive to the rationale' complement.
For any specific $g$, we are finding a rationale $R(g)$ as follows:
\begin{gather}\label{equ:ird-invariance}
    f(R(g))=f(g),\quad Y\perp C(G) | R(G).
\end{gather}

\subsection{Connecting GCL and IRD}
Jointly analyzing the invariance views on GCL and IRD (\cf Equations \eqref{equ:gcl-invariance} and \eqref{equ:ird-invariance}), we find the rationale discovery in IRD suitable to address the limitations of graph augmentations in existing GCL frameworks, answering \emph{What augmentations should the representations be invariant to?}.
Specifically, when conducting instance-discrimination in GCL, we could extract the rationale from the anchor graph, which latches on instance-discriminative information.
We then reveal the rationale that the instance-discrimination task should be invariant to in the augmented views, while concealing its complement.

\section{Methodology}

Inspired by IRD, we propose a new pre-training scheme, Rationale-aware Graph Contrastive Learning (RGCL), which discovers rationales to garner instance-discriminative semantics, and then performs contrastive learning on the rationale-aware views.
We next present the pipeline of RGCL as shown in Figure \ref{fig:framework}: rationale-aware graph augmentation, representation learning, and contrastive learning.

\subsection{Rationale-aware Graph Augmentation}

Prior study \cite{GraphCL} has shown that, compared with edge perturbation and attribute adjusting, node dropping benefits downstream tasks across different categories of graph datasets. Thus, given an anchor graph, we focus on identifying a subset of salient nodes with edges between them as its rationale.

Towards this end, we need to assess the attribution of each node to instance-discrimination.
Moreover, to maintain the diversity of augmented views, it is not sagacious to model the node contribution and rationale generation in a deterministic manner.
Hence, we adopt the idea of probabilistic sampling upon an approximation: given an anchor graph $G=g$, its rationale $R(G)$ follows a probability distribution $P_{R}(R(G)|G=g)$, which summarizes the probability of each node being salient:
\begin{align}\label{equ:rationale-distribution}
    P_R(R(G)&=R(g)|G=g)\nonumber\\
    &=\prod_{v \in \Set{V}_{R}}p(v|g) \prod_{v \in \Set{V}_{C}}(1-p(v|g)),
\end{align}
where $\Set{V}$ and $\Set{V}_{R}$ are the node sets of $g$ and its rationale $R(g)$, respectively; $\Set{V}_{C}=\Set{V}\setminus\Set{V}_{R}$ is the node sets of the complement $C(g)$;
$p(v|g)$ denotes the probability of $v$ being included into $R(g)$, reflecting how semantically important it is. Analogously, we can define the distribution of the rationale's complement $C(G)$ as:
\begin{align}\label{equ:complement-distribution}
    P_C(C(G)&=C(g)|G = g)\nonumber\\
    &=\prod_{v\in \Set{V}_{C}}(1-p(v|g)) \prod_{v\in \Set{V}_{R}}p(v|g),
\end{align}
where $1-p(v|g)$ measures how deficient node $v$ is to accomplish instance-discrimination.

We now present how to implement the foregoing process.
Here we hire a rationale generator network $r(\cdot)$ to parameterize the probability distribution function $p(\cdot|g)$:
\begin{gather}
    \Mat{P}=r(g),
\end{gather}
where $r(\cdot)$ is a GNN-MLP combined encoder that takes the anchor graph $g$ as input and yields normalized node attribution scores $\Mat{P}\in\Space{R}^{|\Set{V}|\times 1}$, where the $v$-th element in $\Mat{P}$ corresponds to aforementioned probability $p(v|g)$.

Further, we sample rationale-aware views from distribution $P_{R}(\cdot| G=g)$ to obtain rationale-aware views:
\begin{gather} \label{equ:rationale-sampling}
    R(g) \sim P_{R}(\cdot| G=g) \quad
    \text{s.t.}\ \ |\Set{V}_{R}| = \rho\cdot|\Set{V}|,
\end{gather}
where based on normalized nodes' attribution scores $\Mat{P}$, we sample $\rho\cdot|\Set{V}_{g}|$ nodes from the original graph $g$, while keeping the edges between sampled nodes.
Similarly, rationale complement views are obtained as:
\begin{gather} \label{equ:complement-sampling}
    C(g) \sim P_{C}(\cdot| G=g) \quad
    \text{s.t.}\ \ |\Set{V}_{C}| = \rho\cdot|\Set{V}|.
\end{gather}

Note that $C(g)$ is not the node difference between $g$ and $R(g)$, but a stochastic complement sampled from the distribution $P_C(\cdot| G=g)$.
As a result, each node within $R(g)$ (or $C(g)$) is assigned with its probability $p(v|g)$ (or $1-p(v|g)$), which illustrates how crucial (or trivial) it is to conduct instance-discrimination in GCL.

\subsection{Rationale-aware Representation Learning}

After sampling from these two distributions, we have $R(g)$ and $C(g)$, while discarding the remaining nodes.
For the rationale-augmented view $R(g)$, we then associate it with the attribution vector $\Mat{P}_{R}\in\Space{R}^{|\Set{V}_{R}|\times 1}$, where we keep node attribute scores in $\Mat{P}$ corresponds to the node set of $R(g)$. 
The complement view $C(g)$ is processed similarly with the attribution vector $\Mat{P}_{C}\in\Space{R}^{|\Set{V}_{C}|\times 1}$.

Having established the rationale-augmented view $R(g)$ with the probability vector $\Mat{P}_{R}$, we feed them into the GNN backbone $f(\cdot)$ (\ie the target model being pre-trained) to generate the rationale-aware representation:
\begin{gather}\label{equ:rationale-representation}
    \Mat{x}_{R}=f(R(g))=\text{Pooling}(\text{GNN}(R(g))\odot\Mat{P}_{R}),
\end{gather}
where $f(\cdot)$ is a combination of the base encoder GNN$(\cdot)$ and the pooling layer Pooling$(\cdot)$, and yields the $d'$-dimensional rationale representation $\Mat{x}_R$.
To be more specific, GNN$(\cdot)$ outputs $\Mat{X}_{R}\in\Space{R}^{|\Set{V}_{R}|\times d'}$ that collects the representations of nodes within $R(g)$.
Subsequently, we apply the element-wise product between $\Mat{X}_{R}$ and $\Mat{P}_{R}$, and then employ the pooling function $\text{Pooling}(\cdot)$ to compress the node representations into the rationale representation.
Note that, when fine-tuning on downstream tasks, we disable the rationale discovery module and discard $\Mat{P}_{R}$ in $f(\cdot)$, \ie $\Mat{x}=\text{Pooling}(\text{GNN}(g))$.

Moreover, we resort to a projection head $h(\cdot)$ to map the graph representation into another latent space where the contrastive learning is conducted, aiming to enforce mutual information between the anchor and rationale to a tighter lower bound \cite{SimCLR}.
Formally, aforementioned process is shown as:
\begin{gather}\label{equ:rationale-projection}
    \Mat{r}=h(\Mat{x}_R)=L_{2}(\text{MLP}(\Mat{x}_R)),
\end{gather}
where $h(\cdot)$ is instantiated by an MLP with $l_{2}$ normalized outputs.
Similarly, we can get the representation and the projection of the complement-aware view $C(g)$:
\begin{gather}\label{equ:complement-representation-projection}
    \Mat{x}_{C}=f(C(g)),\quad \Mat{c}=h(\Mat{x}_{C}).
\end{gather}

\subsection{Rationale-aware Contrastive Learning}

Given the rationale-aware graph augmentations, we move on to achieving the sufficiency and independence principles (\cf Equations \eqref{equ:sufficiency-principle} and \eqref{equ:independence-principle}) in contrastive learning.

Toward this end, for an anchor graph $g$, we randomly sample two rationales from its rationale generation distribution, view them as the positive pair $(R_1(g),R_2(g))$, and establish their projected representations $\Mat{r}_1^+$ and $\Mat{r}_2^+$ via Equation \eqref{equ:rationale-projection}.
Following the previous work \cite{SimCLR}, the rationales of other anchors are treated as the negative views of $g$, and summarized their representations into $\Set{R}^{-}_g$.
We then model the sufficiency principle as minimizing the following contrastive loss:
\begin{gather}\label{equ:loss-su}
    l_{\text{su}}(g)=-\log\frac{\exp{(\Trans{\Mat{r}_1^+}\Mat{r}_2^+/\tau)}}{\sum_{\Mat{r}^-\in\Set{R}^{-}_g}\exp{(\Trans{\Mat{r}_1^+}\Mat{r}^-/\tau)}},
\end{gather}
where $\tau$ is a temperature hyperparameter.
It encourages the agreement between the positive rationale-aware views of the same anchor graph, while enforcing the divergence between different anchors.
Through minimizing $l_{\text{su}}(g)$, rationale generator is required to refine the crucial information in the anchor graph about instance-discrimination, which has a causal relation with semantics (\cf Equations \eqref{equ:sufficiency-principle}).

Going beyond the negative views derived from other anchors' rationales, we further sample one complement $R^c(g)$ from its complement generation distribution, and cast it as the additional negative views of $g$.
And we formalize the independence principle as the minimization of the contrastive loss below:
\begin{gather}\label{equ:loss-in}
    l_{\text{in}}(g)=-\log{\frac{\exp{(\Trans{\Mat{r}_1^+}\Mat{r}_2^+/\tau)}}{\exp{(\Trans{\Mat{r}_1^+}\Mat{r}_2^+/\tau)}+\sum_{\Mat{c}\in\Set{C}}\exp{(\Trans{\Mat{r}_1^+}\Mat{c}/\tau)}}},
\end{gather}
where set $\Set{C}$ summarizes all complement representations appearing in the minibatch data.
The minimization of $l_{\text{in}}(g)$ pushes away the representation of the complement, \ie $\Mat{c}$, from that of the rationale, \ie $\Mat{r}$, making the captured rationale stable regardless of changes of its complement, which consists with the independence principle in Equation \eqref{equ:independence-principle}.

Finally, our objective function conflates these two types of losses, depicting the cooperative game between the rationale generator network $r(\cdot)$ and the target backbone model $f(\cdot)$:
\begin{gather}\label{equ:final_loss_function}
    \min_{r,f,h} \Lapl_{\text{RGCL}}=\Space{E}_{g\in\Set{G}}[l_{\text{su}}(g)+\lambda\cdot l_{\text{in}}(g)],
\end{gather}
where $\lambda$ is the hyperparameter to control the tradeoff between $l_{\text{su}}(g)$ and $l_{\text{in}}(g)$.
After optimization, we throw away the projection head $h(\cdot)$ when fine-tuning on downstream tasks.
It is worth mentioning that our RGCL is a backbone-agnostic self-supervised graph learning framework, which can be applicable to different backbone models. 




\begin{figure*}[t]
\centering
\subcaptionbox{Visualization of Mutag Graphs\label{fig:mutag-vis}}{
	    \vspace{-10pt}
		\includegraphics[width=0.51\textwidth]{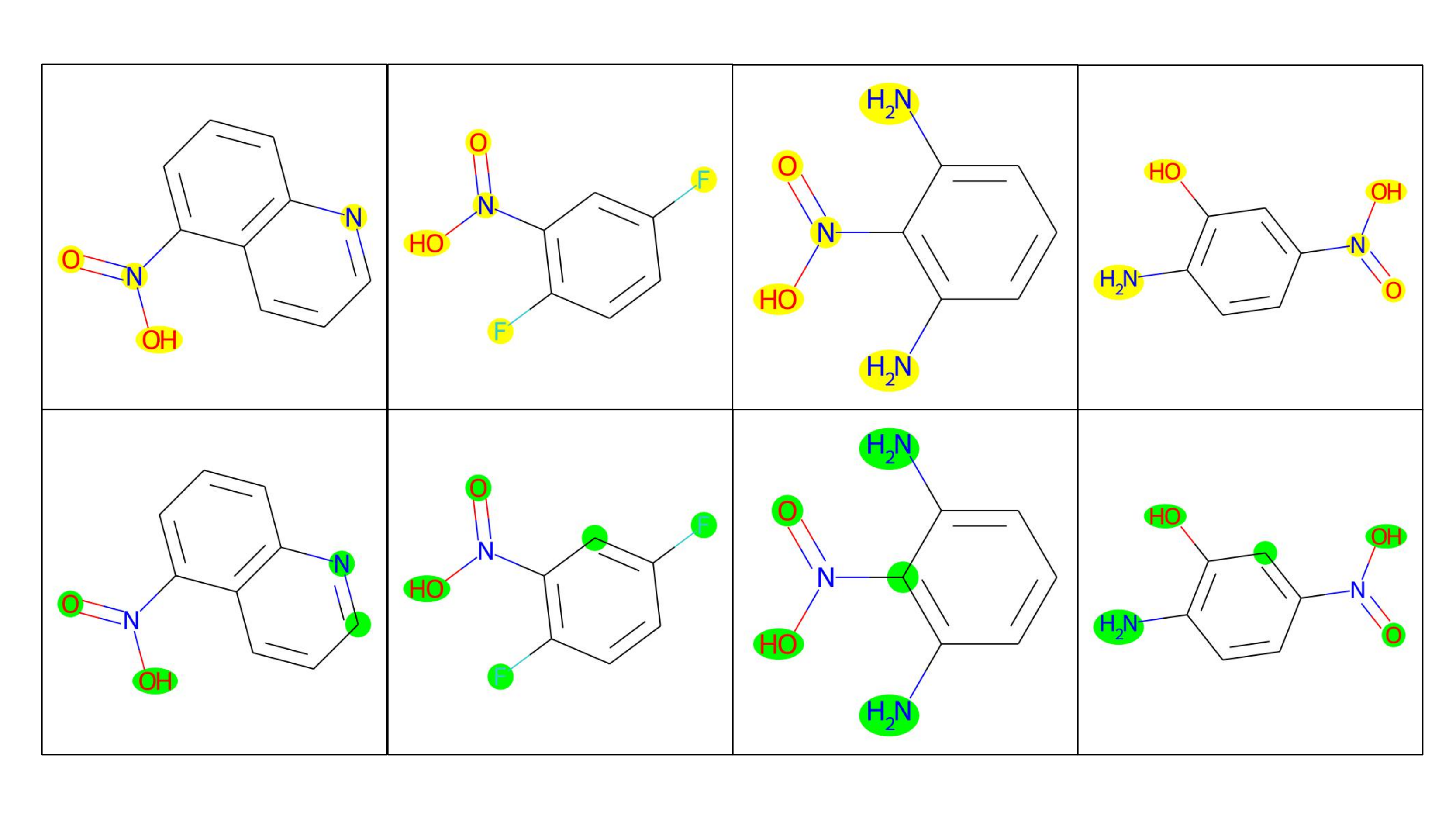}}
\subcaptionbox{Visualization of MNIST-Superpixel Graphs\label{fig:mnist-vis}}{
	    \vspace{-15pt}
		\includegraphics[width=0.45\textwidth]{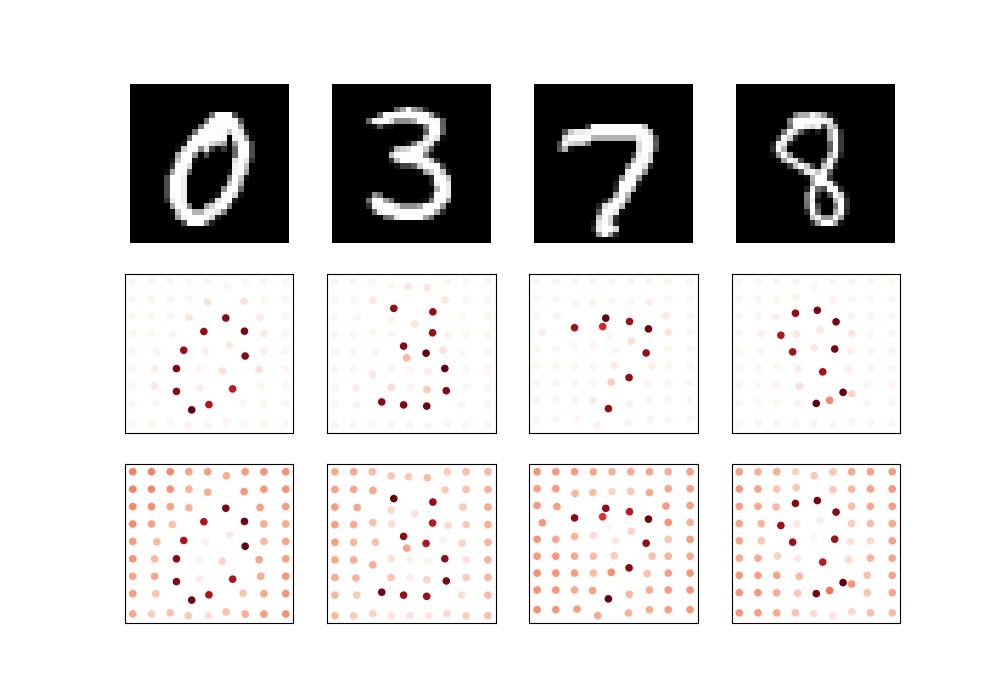}}

\vspace{-5pt}
\caption{(a) The first row presents the rationales labeled by chemistry experts and the second row presents those discovered by our rationale generator. (b) From top to bottom are original MNIST data, MNIST-Superpixel graph and the rationale generated by the rationale generator in RGCL. The darker color indicates higher sampling probability in rationale graphs. }
\label{fig:ra_vis}
\vspace{-10pt}
\end{figure*}

\section{Experiments}

In this section, extensive experiments are conducted to answer two research questions:
\begin{itemize}[leftmargin=*]
    \item \textbf{RQ1:} How effective is the rationale generator in capturing semantically important nodes to construct contrastive views invariant to augmentations? 
    \item \textbf{RQ2:} Whether rationale-aware views promote GCL, further improving the performance of pre-trained backbone model on downstream tasks?
\end{itemize}

\subsection{Effectiveness of Rationale Generator (\textbf{RQ1})}

To justify the effectiveness of our rationale generator, we conducted experiments on two datasets from different domains.

\textbf{Real world biochemical molecule dataset.} 
Following the settings of previous works \cite{GNN-pretrain, GraphCL}, we use Zinc-2M -- 2 million unlabeled molecule graphs sampled from the ZINC15 database \cite{Zinc} to pre-train the backbone model and rationale generator. 
Then the pre-trained rationale generator is applied on a real world molecule benchmark dataset -- Mutag. 
Figure \ref{fig:mutag-vis} presents the rationales labeled by two chemistry experts (the first row in yellow) and those discovered by our RGCL (the second row in green), to interpret ``What part of the molecule graphs are critical to the instance-discriminative property?''. 
Note that the rationale discovered by our rationale generator is the probabilities of each node, so we highlight the top-k nodes in green, where $k$ is the number of nodes highlighted by the experts in corresponding graphs.
Treating the expert-driven rationales as the ground truth, the precision of RGCL-discovered rationales is $78.9\%$.
This comparison justifies the reliability of our rationale discovery.
Details of experiment settings are included in Appendix \ref{app: transfer learning settings}.

\textbf{Superpixel dataset.} Following the setting of \citet{GraphCL}, we formulate the pre-training dataset by omitting the labels of the train set in MNIST-Superpixel dataset and utilize it to pre-train our backbone encoder and rationale generator jointly. 
After pre-training, finetuning is performed in a supervised manner with $1\%$ and $10\%$ data samples with labels randomly chosen from the train set. 
Then we randomly pick two cases to visualize the rationale generated by the rationale generator, which is shown in Figure \ref{fig:mnist-vis}. 
And Table \ref{MNIST-table} shows the performance gain compared with multiple strong baselines.
Due to the limitation of space, experiment settings including the briefs of MNIST-Superpixel and baselines are summarized in Appendix \ref{app: MNIST-superpixel settings}.

We have the following observations on MNIST-Superpixel experiment results in Figure \ref{fig:mnist-vis} and Table \ref{MNIST-table}: 

(1) Our rationale generator is capable of discovering semantic nodes (\emph{e.g.} dark nodes in the original superpixel graph) and the generated rationale helps to preserve these semantic nodes in generated rationale-augmented views, thus providing correct guidance for the following contrastive learning process. 
Consequently, RGCL outperforms GraphCL, a typical GCL framework performing augmentations in a random fashion and other baselines.

\begin{table}[t]
\caption{Classification accuracy (\%) comparison on MNIST-Superpixel. Numbers are from \citet{GraphCL}. \textbf{\underline{Bold}} indicates the best performance while \underline{underline} indicates the second best.}
\label{MNIST-table}
\vspace{10pt}
\begin{center}
\begin{small}
\begin{sc}
\begin{tabular}{lccc}
\toprule
Datasets & $1\%$ MNIST & $10\%$ MNIST\\
\midrule
No Pre-train& 60.39\scriptsize{$\pm$ 1.95}& 79.71\scriptsize{$\pm$ 0.65} \\
Aug. & 67.43\scriptsize{$\pm$ 0.36}& 83.99\scriptsize{$\pm$ 2.19} \\
GAE & 57.58\scriptsize{$\pm$ 2.07}& 86.67\scriptsize{$\pm$ 0.93} \\
Infomax& 63.24\scriptsize{$\pm$ 0.78}& 83.34\scriptsize{$\pm$ 0.24} \\
GraphCL& \underline{83.41\scriptsize{$\pm$ 0.33}}& \underline{93.11\scriptsize{$\pm$ 0.17}} \\
RGCL (ours)& \textbf{\underline{83.84\scriptsize{$\pm$ 0.41}}}& \textbf{\underline{94.92\scriptsize{$\pm$ 0.12}}} \\
\bottomrule
\end{tabular}
\end{sc}
\end{small}
\end{center}
\vspace{-20pt}
\end{table}

\begin{table*}[ht]
\caption{Transfer learning ROC-AUC scores (\%) on downstream graph classification tasks compared with state-of-the-art methods where statistics are from \citet{GraphCL} except GraphLoG and AD-GCL, whose results are reproduced on our platform. \textbf{\underline{Bold}} indicates the best performance while \underline{underline} indicates the second best on each dataset.}
\label{transfer-performance-table}
\vspace{-10pt}
\begin{center}
\begin{scriptsize}
\begin{sc}
\resizebox{\textwidth}{!}{
\begin{tabular}{lcccccccc|cc}
\toprule
{\scriptsize Dataset} & {\scriptsize BBBP} & {\scriptsize Tox21} & {\scriptsize ToxCast} & {\scriptsize SIDER} & {\scriptsize ClinTox} & {\scriptsize MUV} & {\scriptsize HIV} & {\scriptsize BACE} & {\scriptsize AVG.} & {\scriptsize GAIN}\\
\midrule
No Pre-Train& 65.8\tiny{$\pm$4.5}& 74.0\tiny{$\pm$0.8} & 63.4\tiny{$\pm$0.6} & 57.3\tiny{$\pm$1.6} & 58.0\tiny{$\pm$4.4} &71.8 \tiny{$\pm$2.5} & 75.3\tiny{$\pm$1.9} & 70.1\tiny{$\pm$5.4} & 67.0& -\\
AttrMasking& 64.3\tiny{$\pm$ 2.8}& \textbf{\underline{76.7\tiny{$\pm$ 0.4}}}& \textbf{\underline{64.2\tiny{$\pm$ 0.5}}}& \underline{61.0\tiny{$\pm$0.7}} & 71.8\tiny{$\pm$4.1} &74.7\tiny{$\pm$1.4} & 77.2\tiny{$\pm$1.1} &79.3\tiny{$\pm$1.6} &71.1 &4.1\\
ContextPred& 68.0\tiny{$\pm$2.0}& \underline{75.7\tiny{$\pm$0.7}}& \underline{63.9\tiny{$\pm$0.6}} &60.9\tiny{$\pm$0.6} & 65.9\tiny{$\pm$3.8}&\underline{75.8\tiny{$\pm$1.7}} &77.3\tiny{$\pm$1.0} &\underline{79.6\tiny{$\pm$1.2}} &70.9 &3.9\\
GraphCL& 69.68\tiny{$\pm$0.67}& 73.87\tiny{$\pm$0.66}& 62.40\tiny{$\pm$0.57}& 60.53\tiny{$\pm$0.88}& 75.99\tiny{$\pm$2.65}& 69.80\tiny{$\pm$2.66}&\textbf{\underline{78.47\tiny{$\pm$1.22}}} &75.38\tiny{$\pm$1.44} & 70.77&3.77\\
GraphLoG*& \underline{71.04\tiny{$\pm$1.86}}& 74.65\tiny{$\pm$0.60}& 62.32\tiny{$\pm$0.51} & 57.86\tiny{$\pm$1.44}&\underline{78.72\tiny{$\pm$2.58}} & 74.95\tiny{$\pm$1.96}&75.12\tiny{$\pm$1.98} & \textbf{\underline{82.6\tiny{$\pm$1.25}}}& 72.16&5.16\\
AD-GCL*& 68.26\tiny{$\pm$1.03}& 73.56\tiny{$\pm$0.72}& 63.10\tiny{$\pm$0.66} & 59.24\tiny{$\pm$0.86}& 77.63\tiny{$\pm$4.21}& 74.94\tiny{$\pm$2.54}& 75.45\tiny{$\pm$1.28}& 75.02\tiny{$\pm$1.88}& 70.90&3.90\\
RGCL (Ours)& \textbf{\underline{71.42\tiny{$\pm$0.66}}}& 75.20\tiny{$\pm$0.34}& 63.33\tiny{$\pm$0.17} & \textbf{\underline{61.38\tiny{$\pm$0.61}}}& \textbf{\underline{83.38\tiny{$\pm$0.91}}}& \textbf{\underline{76.66\tiny{$\pm$0.99}}}& \underline{77.90\tiny{$\pm$0.80}}& 76.03\tiny{$\pm$0.77}& 73.16&6.16\\
\bottomrule
\end{tabular}}
\end{sc}
\end{scriptsize}
\end{center}
\vspace{-15pt}
\end{table*}

\begin{table*}[ht]
\caption{Unsupervised representation learning classification accuracy (\%) on TU datasets. The compared numbers are from \citet{GraphCL} except AD-GCL, whose statistics are reproduced on our platform. \textbf{\underline{Bold}} indicates the best performance while \underline{underline} indicates the second best on each dataset.}
\label{unsupervised performance table}
\begin{center}
\begin{scriptsize}
\begin{sc}
\begin{tabular}{lcccccccc|cc}
\toprule
{\scriptsize Dataset} & {\scriptsize NCI1} & {\scriptsize PROTEINS} & {\scriptsize DD} & {\scriptsize MUTAG} & {\scriptsize COLLAB} & {\scriptsize RDT-B} & {\scriptsize RDT-M5K} & {\scriptsize IMDB-B} & {\scriptsize AVG.}\\
\midrule
No Pre-Train & 65.40\tiny{$\pm$0.17}& 72.73\tiny{$\pm$0.51} & 75.67.4\tiny{$\pm$0.29} & 87.39\tiny{$\pm$1.09} & 65.29\tiny{$\pm$0.16} &76.86 \tiny{$\pm$0.25} & 48.48\tiny{$\pm$0.28} & 69.37\tiny{$\pm$0.37} & 70.15\\
InfoGraph& 76.20\tiny{$\pm$ 1.06}& \underline{74.44\tiny{$\pm$ 0.31}}& 72.85\tiny{$\pm$ 1.78}& \textbf{\underline{89.01\tiny{$\pm$1.13}}} & 70.05\tiny{$\pm$1.13} &82.50\tiny{$\pm$1.42} & 53.46\tiny{$\pm$1.03} &\textbf{\underline{73.03\tiny{$\pm$0.87}}} &74.02\\
GraphCL& \underline{77.87\tiny{$\pm$0.41}}& 74.39\tiny{$\pm$0.45}& \underline{78.62\tiny{$\pm$0.40}} &86.80\tiny{$\pm$1.34} & \underline{71.36\tiny{$\pm$1.15}}&89.53\tiny{$\pm$0.84} &\underline{55.99\tiny{$\pm$0.28}} &71.14\tiny{$\pm$0.44} &75.71\\
AD-GCL& 73.91\tiny{$\pm$0.77}& 73.28\tiny{$\pm$0.46}& 75.79\tiny{$\pm$0.87}& \underline{88.74\tiny{$\pm$1.85}}& \textbf{\underline{72.02\tiny{$\pm$0.56}}}& \underline{90.07\tiny{$\pm$0.85}}&54.33\tiny{$\pm$0.32} &70.21\tiny{$\pm$0.68} & 74.79\\
Ours& \textbf{\underline{78.14\tiny{$\pm$1.08}}}& \textbf{\underline{75.03\tiny{$\pm$0.43}}}& \textbf{\underline{78.86\tiny{$\pm$0.48}}} & 87.66\tiny{$\pm$1.01}& 70.92\tiny{$\pm$0.65}& \textbf{\underline{90.34\tiny{$\pm$0.58}}}& \textbf{\underline{56.38\tiny{$\pm$0.40}}}& \underline{71.85\tiny{$\pm$0.84}}& 76.15\\
\bottomrule
\end{tabular}
\end{sc}
\end{scriptsize}
\end{center}
\vspace{-15pt}
\end{table*}

(2) The nodes' color distribution in the visualized rationale graph follows a dark-light-dark pattern from center to edge, which attracts our attention and an intuitive explanation is provided as follows.
Semantic nodes in the center of the graph are captured by the rationale generator and assigned the highest probability, so we call this zone semantic zone. 
However, sampling nodes close to semantic ones when generating augmented views could possibly lead to semantic information change, for example, adding a node in the center of a \textbf{\emph{0}} superpixel graph may change the semantic information from \textbf{\emph{0}} to \textbf{\emph{8}} and adding a node at the up-left part of a \textbf{\emph{1}} superpixel graph makes semantic information change from \textbf{\emph{1}} to \textbf{\emph{7}}. 
Therefore we name this zone near the semantic zone as confusion zone and nodes in this area are supposed to be assigned the lowest sampling probability in order to avoid semantic information change. 
Nodes at the edge part of the graph contain no semantic information but including some of them in the augmented views benefits the diversity of the augmented views, making the backbone model more robust. 
Hence, these nodes are supposed to be sampled with moderate probabilities and the edge part is named as background zone. 
During the contrastive loss optimization process, the rationale generator is trained to discover the semantic nodes and push away data samples from each other to prevent confusion. 
Consequently, from the center part to edge part, the semantic zone - confusion zone - background zone follows a dark-light-dark pattern.

\textbf{Summary.} The overall results of these two experiments answered \textbf{RQ1} and demonstrated the effectiveness of our proposed rationale generator.

\subsection{Performance on Downstream Tasks (\textbf{RQ2})}

\begin{table*}[ht]
\caption{Ablation study for RGCL on downstream transfer learning datasets.}
\label{ablation-table}
\begin{center}
\begin{scriptsize}
\begin{sc}
\begin{tabular}{lcccccccc|c}
\toprule
{\scriptsize Dataset} & {\scriptsize BBBP} & {\scriptsize Tox21} & {\scriptsize ToxCast} & {\scriptsize SIDER} & {\scriptsize ClinTox} & {\scriptsize MUV} & {\scriptsize HIV} & {\scriptsize BACE} & {\scriptsize AVG.} \\
\midrule

RGCL w/o RV& 70.08\tiny{$\pm$0.86}& 74.01\tiny{$\pm$0.56}& 62.23\tiny{$\pm$0.38} & 60.38\tiny{$\pm$0.52}& 77.29\tiny{$\pm$1.81}& 70.12\tiny{$\pm$1.28}& 77.42\tiny{$\pm$0.94}& 75.08\tiny{$\pm$1.22}& 70.83\\
RGCL w/o I& 70.64\tiny{$\pm$0.92}& 75.02\tiny{$\pm$0.48}& 63.12\tiny{$\pm$0.56} & 61.20\tiny{$\pm$0.56}& 84.26\tiny{$\pm$1.55}& 73.64\tiny{$\pm$1.08}& 77.08\tiny{$\pm$1.21}& 75.88\tiny{$\pm$0.94}& 72.60\\
RGCL& 71.42\tiny{$\pm$0.66}& 75.20\tiny{$\pm$0.34}& 63.33\tiny{$\pm$0.17} & 61.38\tiny{$\pm$0.61}& 83.38\tiny{$\pm$0.91}& 76.66\tiny{$\pm$0.99}& 77.90\tiny{$\pm$0.80}& 76.03\tiny{$\pm$0.77}& 73.16\\
\bottomrule
\end{tabular}
\end{sc}
\end{scriptsize}
\end{center}
\vspace{-15pt}
\end{table*}

\textbf{Transfer learning.} We first pre-train a backbone model on Zinc-2M \cite{Zinc} and then finetune it on 8 benchmark multi-task binary classification datasets in biochemistry domain, which are contained in MoleculeNet \cite{wu2018moleculenet}. 
Note that the downstream datasets are splited using scaffold split in order to simulate the real world case --- out-of-distribution --- and examine the generalization ability of the pre-trained model.
Details of experiment settings are included in Appendix \ref{app: transfer learning settings}.

We adopt AttrMasking \cite{GNN-pretrain}, ContextPred \cite{GNN-pretrain}, GraphCL \cite{GraphCL}, GraphLoG \cite{graph-l-o-g} and AD-GCL \cite{AD-GCL}, which are state-of-the-art pre-training paradigms in this area, as our baselines. 
Finetuning procedure is repeated for 10 times with different random seeds and we evaluate the mean and standard deviation of ROC-AUC scores on each downstream dataset, which is consistent with our baselines. 
Further, the average and its gain compared with the training-from-scratch model are listed as well. 
All transfer learning performance on downstream tasks are presented in Table \ref{transfer-performance-table}.

With the guidance of rationale to construct semantic information preserved views, our RGCL framework achieves best performance on 4 out of 8 datasets and highest average gain compared with existing baselines. 
Compared with GraphCL \cite{GraphCL}, our framework RGCL leverages an rationale generator to construct rationale-aware views and a mixture loss function in Equation \eqref{equ:final_loss_function} to ensure both sufficiency and independency for generated rationales. 
So we go a step further to compare the performance of GraphCL with RGCL to demonstrate the crucial role of rationale-augmented views in GCL. 
Rationale-augmented views in RGCL preserve more semantic information in anchor graphs, thus making it outperform GraphCL in 7 out of 8 downstream datasets except HIV, and raising the average performance gain from $3.77$ to $6.16$. 
As for the performance deterioration on HIV dataset, one possible reason is that the rationale probability distribution in HIV is different from that in the pre-training dataset ZINC15, making the pre-trained model fail to capture discriminative features in HIV dataset. 
Hence it is challenging to train a robust rationale generator with great generalization ability, which we will explore in the future.

\textbf{Unsupervised learning.} Furthermore, we follow the settings of InfoGraph \cite{infograph} to evaluate RGCL in the unsupervised representation learning where the encoded graph representations are fed into a non-linear SVM classifier.
We compare RGCL with four self-supervised learning baselines on TU datasets \cite{TU-dataset}: untrained GIN \cite{GIN}, InfoGraph \cite{infograph}, GraphCL \cite{GraphCL} and AD-GCL \cite{AD-GCL} with the default settings in \citet{infograph}. 
Summaries of datasets and baselines are provided in Appendix \ref{app: unsupervised learning settings}. 
Table \ref{unsupervised performance table} shows that RGCL outperforms the baselines on 5 out of 8 cases and achieves the best average performance.

\textbf{Summary.} Extensive experiments on transfer learning and unsupervised representation learning demonstrate the state-of-the-art performance of RGCL, answering \textbf{RQ2}: rationale-aware views benefit GCL, endowing the pre-trained model with better transferability and generalization ability.

\subsection{Ablation Study}

\begin{figure}[t]
    \begin{center}
    \centerline{\includegraphics[width=\columnwidth]{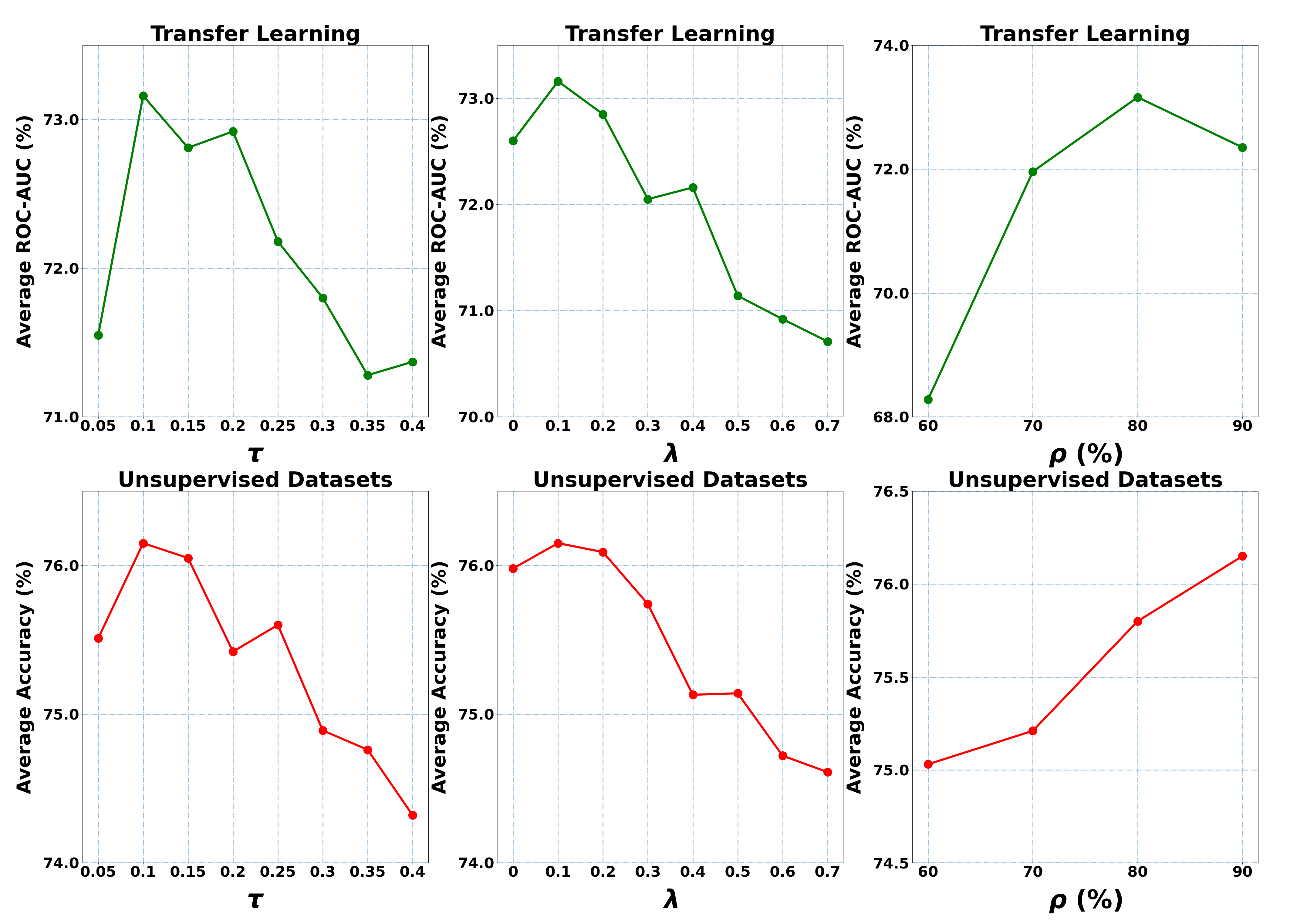}}
    \vspace{-10pt}
    \caption{Sensitivity \wrt hyperparameter $\tau$, $\lambda$ and $\rho$.}
    \label{fig:ab_exp}
    \end{center}
    \vspace{-30pt}
\end{figure}

In this section, ablation studies are conducted to demonstrate the importance of rationale-views and independence term $l_{in}$ in loss function (\cf Equation \eqref{equ:final_loss_function}). 
We first replace our rationale-aware data augmentation module with one in a random fashion (\ie randomly discarding certain nodes to construct augmented views). 
As illustrated in Table \ref{ablation-table}, the performance of this new GCL framework without rationale views (RGCL w/o RV), decreases sharply compared with RGCL, demonstrating the crucial role of rationale-aware views in guiding contrastive learner. 
Note that RGCL w/o RV can be viewed as GraphCL \cite{GraphCL} but a pooling layer with attention mechanism. 
Then we ablate the final loss function (\emph{e.g.} eliminating independence term $l_{in}$ in it). 
The performance drop of RGCL without independence term (RGCL w/o I) demonstrates the strength of independence principle in IRD, which is regulating the information outside the rationale.      

\subsection{Hyperparameter sensitivity}

RGCL's sensitivity \emph{w.r.t.} hyperparameters $\tau$, $\lambda$ and $\rho$ is evaluated by presenting the average performances on both transfer learning and unsupervised settings as in Figure \ref{fig:ab_exp}. And we have following observations:

\begin{itemize}[leftmargin=*]
    \item As verified in \citet{temperature-in-CL}, $\tau$ plays a crucial role in contrastive learning. A too small value of $\tau$ (\emph{e.g.}, $0.05$) hurts RGCL's performance; continuously increasing its value (\ie over $0.2$) leads to performance drop as well. Thus we suggest to tune $\tau$ around 0.1 carefully on both MoleculeNet \cite{wu2018moleculenet} and TU \cite{TU-dataset} datasets. Moreover, according to our results, different datasets may desire different settings of $\tau$.

    \item The independence term $l_{in}$ works as a regulizer in optimization process, so the performance drops when overemphasizing it (\ie setting a large value for $\lambda$). Nevertheless, a proper setting of regulizer (\ie $\lambda=0.1$) prevents the model from degeneration, benefiting contrastive learning. 
    
    \item Hyperparameter $\rho$ controls the scale of the augmented views. We suggest to tune it around a comparatively large value (\eg $80\%$), because a too small value enlarges the distribution gap between augmented views (\ie graphs of small scales) and anchor graphs (\ie graphs of original scales), undermining the representative ability of pre-trained model on downstream tasks.  
    
\end{itemize}

\section{Conclusion}

In this paper, we revisit graph constrastive learning (GCL) from the standpoint of invariant rationale discovery (IRD) and propose a novel graph contrastive learning framework, \underline{R}ationale-aware \underline{G}raph \underline{C}ontrastive \underline{L}earning (RGCL) for GNN pre-training. 
In our framework, a GNN-MLP combined rationale generator requires no domain knowledge nor manually designed features to generate rationales - discovering semantic information related nodes in graph instances. 
Under the correct guidance of the rationale, semantic information preserved views are generated to conduct contrastive learning. 
Visual inspections on MNIST-Superpixel dataset demonstrate the effectiveness of rationale generator by revealing its capability of discovering discriminative nodes. 
Extensive experiments conducted on biochemical molecule and social network benchmark datasets show that rationale-aware views promote pre-trained models' transferability and generalization ability, thus achieving state-of-the-art performance compared with baselines.

In future, we will explore the fine-tuning of rationale generator on downstream tasks, such that the pre-training scheme in GCL endows the rationales with a local view of self-discriminative patterns and the fine-tuning will enhance the rationales' global view of class-wise discriminative patterns.
It is promising to probe into explainability of GCL.
Moreover, we would like to investigate the retrospective and introspective learning of rationale discovery, which in turn guide the discrimination tasks and improve the generalization ability of the backbone models \cite{JTT}.

\section{Acknowledgements}
This work is supported by the National Key Research and Development Program of China (2021ZD0111802) and National Natural Science Foundation of China (U19A2079, U21B2026, U1936210). This research is also supported by CCCD Key Lab of Ministry of Culture and Tourism and Sea-NExT Joint Lab.

\bibliography{myicml2022}

\begin{thebibliography}{43}
\providecommand{\natexlab}[1]{#1}
\providecommand{\url}[1]{\texttt{#1}}
\expandafter\ifx\csname urlstyle\endcsname\relax
  \providecommand{\doi}[1]{doi: #1}\else
  \providecommand{\doi}{doi: \begingroup \urlstyle{rm}\Url}\fi

\bibitem[Brown et~al.(2020)Brown, Mann, Ryder, Subbiah, Kaplan, Dhariwal,
  Neelakantan, Shyam, Sastry, Askell, Agarwal, Herbert{-}Voss, Krueger,
  Henighan, Child, Ramesh, Ziegler, Wu, Winter, Hesse, Chen, Sigler, Litwin,
  Gray, Chess, Clark, Berner, McCandlish, Radford, Sutskever, and
  Amodei]{GPT-3}
Brown, T.~B., Mann, B., Ryder, N., Subbiah, M., Kaplan, J., Dhariwal, P.,
  Neelakantan, A., Shyam, P., Sastry, G., Askell, A., Agarwal, S.,
  Herbert{-}Voss, A., Krueger, G., Henighan, T., Child, R., Ramesh, A.,
  Ziegler, D.~M., Wu, J., Winter, C., Hesse, C., Chen, M., Sigler, E., Litwin,
  M., Gray, S., Chess, B., Clark, J., Berner, C., McCandlish, S., Radford, A.,
  Sutskever, I., and Amodei, D.
\newblock Language models are few-shot learners.
\newblock In \emph{NeurIPS}, 2020.

\bibitem[Chang et~al.(2020)Chang, Zhang, Yu, and Jaakkola]{InvRationale}
Chang, S., Zhang, Y., Yu, M., and Jaakkola, T.~S.
\newblock Invariant rationalization.
\newblock In \emph{{ICML}}, volume 119, pp.\  1448--1458, 2020.

\bibitem[Chen et~al.(2018)Chen, Song, Wainwright, and Jordan]{L2X}
Chen, J., Song, L., Wainwright, M.~J., and Jordan, M.~I.
\newblock Learning to explain: An information-theoretic perspective on model
  interpretation.
\newblock In \emph{{ICML}}, volume~80, pp.\  882--891, 2018.

\bibitem[Chen et~al.(2020)Chen, Kornblith, Norouzi, and Hinton]{SimCLR}
Chen, T., Kornblith, S., Norouzi, M., and Hinton, G.~E.
\newblock A simple framework for contrastive learning of visual
  representations.
\newblock In \emph{{ICML}}, volume 119, pp.\  1597--1607, 2020.

\bibitem[Dangovski et~al.(2021)Dangovski, Jing, Loh, Han, Srivastava, Cheung,
  Agrawal, and Soljacic]{ECL}
Dangovski, R., Jing, L., Loh, C., Han, S., Srivastava, A., Cheung, B., Agrawal,
  P., and Soljacic, M.
\newblock Equivariant contrastive learning.
\newblock \emph{CoRR}, abs/2111.00899, 2021.

\bibitem[Devlin et~al.(2018)Devlin, Chang, Lee, and Toutanova]{BERT}
Devlin, J., Chang, M., Lee, K., and Toutanova, K.
\newblock {BERT:} pre-training of deep bidirectional transformers for language
  understanding.
\newblock \emph{CoRR}, abs/1810.04805, 2018.

\bibitem[Dwivedi et~al.(2020)Dwivedi, Joshi, Laurent, Bengio, and
  Bresson]{Benchmarking}
Dwivedi, V.~P., Joshi, C.~K., Laurent, T., Bengio, Y., and Bresson, X.
\newblock Benchmarking graph neural networks.
\newblock \emph{CoRR}, 2020.

\bibitem[Goyal et~al.(2019)Goyal, Mahajan, Gupta, and Misra]{benchmark-ssl-cv}
Goyal, P., Mahajan, D., Gupta, A., and Misra, I.
\newblock Scaling and benchmarking self-supervised visual representation
  learning.
\newblock In \emph{ICCV}, pp.\  6391--6400, 2019.

\bibitem[Hamilton et~al.(2017)Hamilton, Ying, and Leskovec]{GraphSAGE}
Hamilton, W., Ying, Z., and Leskovec, J.
\newblock Inductive representation learning on large graphs.
\newblock \emph{NeurIPS}, 30, 2017.

\bibitem[He et~al.(2020)He, Fan, Wu, Xie, and Girshick]{he2020momentum}
He, K., Fan, H., Wu, Y., Xie, S., and Girshick, R.
\newblock Momentum contrast for unsupervised visual representation learning.
\newblock In \emph{CVPR}, pp.\  9729--9738, 2020.

\bibitem[Hu et~al.(2020)Hu, Liu, Gomes, Zitnik, Liang, Pande, and
  Leskovec]{GNN-pretrain}
Hu, W., Liu, B., Gomes, J., Zitnik, M., Liang, P., Pande, V.~S., and Leskovec,
  J.
\newblock Strategies for pre-training graph neural networks.
\newblock In \emph{{ICLR}}, 2020.

\bibitem[Kipf \& Welling(2016)Kipf and Welling]{GAE}
Kipf, T.~N. and Welling, M.
\newblock Variational graph auto-encoders.
\newblock \emph{CoRR}, abs/1611.07308, 2016.

\bibitem[Kipf \& Welling(2017)Kipf and Welling]{GCN}
Kipf, T.~N. and Welling, M.
\newblock Semi-supervised classification with graph convolutional networks.
\newblock In \emph{{ICLR} (Poster)}. OpenReview.net, 2017.

\bibitem[Kolesnikov et~al.(2019)Kolesnikov, Zhai, and Beyer]{revisit-ssl-cv}
Kolesnikov, A., Zhai, X., and Beyer, L.
\newblock Revisiting self-supervised visual representation learning.
\newblock In \emph{CVPR}, pp.\  1920--1929, 2019.

\bibitem[Li et~al.(2022)Li, Wang, Xiao, Ji, and seng Chua]{IGV}
Li, Y., Wang, X., Xiao, J., Ji, W., and seng Chua, T.
\newblock Invariant grounding for video question answering.
\newblock In \emph{CVPR}, 2022.

\bibitem[Liu et~al.(2021{\natexlab{a}})Liu, Haghgoo, Chen, Raghunathan, Koh,
  Sagawa, Liang, and Finn]{JTT}
Liu, E.~Z., Haghgoo, B., Chen, A.~S., Raghunathan, A., Koh, P.~W., Sagawa, S.,
  Liang, P., and Finn, C.
\newblock Just train twice: Improving group robustness without training group
  information.
\newblock In \emph{{ICML}}, volume 139, pp.\  6781--6792, 2021{\natexlab{a}}.

\bibitem[Liu et~al.(2021{\natexlab{b}})Liu, Wang, Liu, Lasenby, Guo, and
  Tang]{GraphMVP}
Liu, S., Wang, H., Liu, W., Lasenby, J., Guo, H., and Tang, J.
\newblock Pre-training molecular graph representation with 3d geometry.
\newblock \emph{CoRR}, abs/2110.07728, 2021{\natexlab{b}}.

\bibitem[Misra \& van~der Maaten(2020)Misra and van~der Maaten]{PIRL}
Misra, I. and van~der Maaten, L.
\newblock Self-supervised learning of pretext-invariant representations.
\newblock In \emph{{CVPR}}, pp.\  6706--6716, 2020.

\bibitem[Monti et~al.(2017)Monti, Boscaini, Masci, Rodola, Svoboda, and
  Bronstein]{MNIST-dataset}
Monti, F., Boscaini, D., Masci, J., Rodola, E., Svoboda, J., and Bronstein,
  M.~M.
\newblock Geometric deep learning on graphs and manifolds using mixture model
  cnns.
\newblock In \emph{CVPR}, pp.\  5115--5124, 2017.

\bibitem[Morris et~al.(2020)Morris, Kriege, Bause, Kersting, Mutzel, and
  Neumann]{TU-dataset}
Morris, C., Kriege, N.~M., Bause, F., Kersting, K., Mutzel, P., and Neumann, M.
\newblock Tudataset: {A} collection of benchmark datasets for learning with
  graphs.
\newblock \emph{CoRR}, abs/2007.08663, 2020.

\bibitem[Qiu et~al.(2020)Qiu, Chen, Dong, Zhang, Yang, Ding, Wang, and
  Tang]{GCC}
Qiu, J., Chen, Q., Dong, Y., Zhang, J., Yang, H., Ding, M., Wang, K., and Tang,
  J.
\newblock Gcc: Graph contrastive coding for graph neural network pre-training.
\newblock In \emph{SIGKDD}, pp.\  1150--1160, 2020.

\bibitem[Rong et~al.(2020)Rong, Bian, Xu, Xie, Wei, Huang, and Huang]{GROVER}
Rong, Y., Bian, Y., Xu, T., Xie, W., Wei, Y., Huang, W., and Huang, J.
\newblock Self-supervised graph transformer on large-scale molecular data.
\newblock In \emph{NeurIPS}, 2020.

\bibitem[Sterling \& Irwin(2015)Sterling and Irwin]{Zinc}
Sterling, T. and Irwin, J.~J.
\newblock Zinc 15--ligand discovery for everyone.
\newblock \emph{Journal of chemical information and modeling}, 55\penalty0
  (11):\penalty0 2324--2337, 2015.

\bibitem[Sun et~al.(2020)Sun, Hoffmann, Verma, and Tang]{infograph}
Sun, F., Hoffmann, J., Verma, V., and Tang, J.
\newblock Infograph: Unsupervised and semi-supervised graph-level
  representation learning via mutual information maximization.
\newblock In \emph{{ICLR}}, 2020.

\bibitem[Suresh et~al.(2021)Suresh, Li, Hao, and Neville]{AD-GCL}
Suresh, S., Li, P., Hao, C., and Neville, J.
\newblock Adversarial graph augmentation to improve graph contrastive learning.
\newblock \emph{CoRR}, abs/2106.05819, 2021.

\bibitem[van~den Oord et~al.(2018)van~den Oord, Li, and Vinyals]{infonce}
van~den Oord, A., Li, Y., and Vinyals, O.
\newblock Representation learning with contrastive predictive coding.
\newblock \emph{CoRR}, abs/1807.03748, 2018.

\bibitem[Velickovic et~al.(2018)Velickovic, Fedus, Hamilton, Li{\`{o}}, Bengio,
  and Hjelm]{DGI}
Velickovic, P., Fedus, W., Hamilton, W.~L., Li{\`{o}}, P., Bengio, Y., and
  Hjelm, R.~D.
\newblock Deep graph infomax.
\newblock \emph{CoRR}, abs/1809.10341, 2018.

\bibitem[Wang \& Liu(2021)Wang and Liu]{temperature-in-CL}
Wang, F. and Liu, H.
\newblock Understanding the behaviour of contrastive loss.
\newblock In \emph{{CVPR}}, pp.\  2495--2504, 2021.

\bibitem[Wang et~al.(2021{\natexlab{a}})Wang, Wu, Zhang, He, and seng
  Chua]{ReFine}
Wang, X., Wu, Y., Zhang, A., He, X., and seng Chua, T.
\newblock Towards multi-grained explainability for graph neural networks.
\newblock In \emph{NeurIPS}, 2021{\natexlab{a}}.

\bibitem[Wang et~al.(2022)Wang, Wu, Zhang, Feng, He, and Chua]{RC-Explainer}
Wang, X., Wu, Y., Zhang, A., Feng, F., He, X., and Chua, T.-S.
\newblock Reinforced causal explainer for graph neural networks.
\newblock \emph{TPAMI}, 2022.

\bibitem[Wang et~al.(2021{\natexlab{b}})Wang, Min, Chen, and Wu]{wang2021multi}
Wang, Y., Min, Y., Chen, X., and Wu, J.
\newblock Multi-view graph contrastive representation learning for drug-drug
  interaction prediction.
\newblock In \emph{WWW}, pp.\  2921--2933, 2021{\natexlab{b}}.

\bibitem[Wu et~al.(2022)Wu, Wang, Zhang, He, and Chua]{DIR}
Wu, Y., Wang, X., Zhang, A., He, X., and Chua, T.-S.
\newblock Discovering invariant rationales for graph neural networks.
\newblock In \emph{ICLR}, 2022.

\bibitem[Wu et~al.(2017)Wu, Ramsundar, Feinberg, Gomes, Geniesse, Pappu,
  Leswing, and Pande]{Molecule-Net}
Wu, Z., Ramsundar, B., Feinberg, E.~N., Gomes, J., Geniesse, C., Pappu, A.~S.,
  Leswing, K., and Pande, V.~S.
\newblock Moleculenet: {A} benchmark for molecular machine learning.
\newblock \emph{CoRR}, abs/1703.00564, 2017.

\bibitem[Wu et~al.(2018)Wu, Ramsundar, Feinberg, Gomes, Geniesse, Pappu,
  Leswing, and Pande]{wu2018moleculenet}
Wu, Z., Ramsundar, B., Feinberg, E.~N., Gomes, J., Geniesse, C., Pappu, A.~S.,
  Leswing, K., and Pande, V.
\newblock Moleculenet: a benchmark for molecular machine learning.
\newblock \emph{Chemical science}, 9\penalty0 (2):\penalty0 513--530, 2018.

\bibitem[Wu et~al.(2021)Wu, Pan, Chen, Long, Zhang, and Yu]{GNNSurvey}
Wu, Z., Pan, S., Chen, F., Long, G., Zhang, C., and Yu, P.~S.
\newblock A comprehensive survey on graph neural networks.
\newblock \emph{TNNLS}, 32\penalty0 (1):\penalty0 4--24, 2021.

\bibitem[Xu et~al.(2019)Xu, Hu, Leskovec, and Jegelka]{GIN}
Xu, K., Hu, W., Leskovec, J., and Jegelka, S.
\newblock How powerful are graph neural networks?
\newblock In \emph{{ICLR}}, 2019.

\bibitem[Xu et~al.(2021)Xu, Wang, Ni, Guo, and Tang]{graph-l-o-g}
Xu, M., Wang, H., Ni, B., Guo, H., and Tang, J.
\newblock Self-supervised graph-level representation learning with local and
  global structure.
\newblock In \emph{{ICML}}, volume 139, pp.\  11548--11558, 2021.

\bibitem[Ying et~al.(2019)Ying, Bourgeois, You, Zitnik, and
  Leskovec]{GNNExplainer}
Ying, Z., Bourgeois, D., You, J., Zitnik, M., and Leskovec, J.
\newblock Gnnexplainer: Generating explanations for graph neural networks.
\newblock In \emph{NeurIPS}, pp.\  9240--9251, 2019.

\bibitem[You et~al.(2020)You, Chen, Sui, Chen, Wang, and Shen]{GraphCL}
You, Y., Chen, T., Sui, Y., Chen, T., Wang, Z., and Shen, Y.
\newblock Graph contrastive learning with augmentations.
\newblock \emph{NeurIPS}, 33:\penalty0 5812--5823, 2020.

\bibitem[Yu et~al.(2019)Yu, Chang, Zhang, and
  Jaakkola]{DBLP:conf/emnlp/YuCZJ19}
Yu, M., Chang, S., Zhang, Y., and Jaakkola, T.~S.
\newblock Rethinking cooperative rationalization: Introspective extraction and
  complement control.
\newblock In \emph{{EMNLP/IJCNLP}}, pp.\  4092--4101, 2019.

\bibitem[Zhang et~al.(2020)Zhang, Hu, Subramonian, and Sun]{motif-GCL}
Zhang, S., Hu, Z., Subramonian, A., and Sun, Y.
\newblock Motif-driven contrastive learning of graph representations.
\newblock \emph{CoRR}, abs/2012.12533, 2020.

\bibitem[Zhu et~al.(2020)Zhu, Xu, Yu, Liu, Wu, and Wang]{GRACE}
Zhu, Y., Xu, Y., Yu, F., Liu, Q., Wu, S., and Wang, L.
\newblock Deep graph contrastive representation learning.
\newblock \emph{CoRR}, abs/2006.04131, 2020.

\bibitem[Zhu et~al.(2021)Zhu, Xu, Yu, Liu, Wu, and Wang]{GCA}
Zhu, Y., Xu, Y., Yu, F., Liu, Q., Wu, S., and Wang, L.
\newblock Graph contrastive learning with adaptive augmentation.
\newblock In \emph{WWW}, pp.\  2069--2080, 2021.

\end{thebibliography}
\bibliographystyle{icml2022}

\newpage
\appendix
\onecolumn
\section{Rationale-aware Graph Contrastive Learning (RGCL) algorithm}
\begin{algorithm}[ht]
   \caption{RGCL algorithm}
   \label{alg:RGCL}
\begin{algorithmic}
   \STATE {\bfseries Initialize:} dataset \{$g_m: m = 1, 2, ..., M $\}, encoder $f(\cdot)$, rationale generator $r(\cdot)$, projector $h(\cdot)$, sampling ratio $\rho$ and hyperparameter $\lambda$.
   \FOR{sampled minibatch of data \{$g_n: n =1, 2, ..., N $\}}
   \FOR{$n=1$ {\bfseries to} $N$}
   \STATE $\Mat{P}_n=r(g_n)$
   \STATE $R'(g_n) \sim P_{R}(R(G)|G=g_n), \quad R''(g_n) \sim P_{R}(R(G)|G=g_n)$ \kern 2.5pc \# Equation \eqref{equ:rationale-sampling}
   \STATE $R^c(g_n) \sim P_{C}({R^c}(G)|G=g_n)$ \kern 13.5pc \# Equation \eqref{equ:complement-sampling}
   \STATE $\Mat{r}'_n=h(f(R'(g_n)))$, \quad $\Mat{r}''_n=h(f(R''(g_n)))$
   \STATE $\Mat{c}_n=h(f(R^c(g_n)))$
   \ENDFOR
   \FOR{$n=1$ {\bfseries to} $N$}
   \STATE $\Set{R}^{-}_n = \{\Mat{r}'_i, \Mat{r}''_i: i = 1, 2, ..., n-1, n+1, ..., N\}$
   \STATE $\Set{C} = \{\Mat{c}_i: i = 1, 2, ..., N\}$
   \STATE $l_{\text{su}}^n=-\log\frac{\exp{(\Trans{\Mat{r}'_n}\Mat{r}''_n/\tau)}}{\sum_{\Mat{r}^{-}\in\Set{R}^{-}_n}\exp{(\Trans{\Mat{r}'_n}\Mat{r}^{-}/\tau)}}$ \kern 12.5pc \# Equation \eqref{equ:loss-su}
   \STATE $l_{\text{in}}^n=-\log{\frac{\exp{(\Trans{\Mat{r}'_n}\Mat{r}''_n/\tau)}}{\exp{(\Trans{\Mat{r}'_n}\Mat{r}''_n/\tau)}+\sum_{\Mat{c}\in\Set{C}}\exp{(\Trans{\Mat{r}'_n}\Mat{c}/\tau)}}}$ \kern 9.5pc \# Equation \eqref{equ:loss-in}
   \ENDFOR
   \STATE $\Lapl_{\text{RGCL}}=\frac{1}{N}\sum_{n=1}^N(l_{\text{su}}^n+\lambda\cdot l_{\text{in}}^n)$ \kern 14pc \# Equation \eqref{equ:final_loss_function}
   \STATE Update $f(\cdot)$, $r(\cdot)$, and $h(\cdot)$ to minimize $\Lapl_{\text{RGCL}}$
   \ENDFOR
   \STATE {\bfseries return} Encoder $f(\cdot)$
\end{algorithmic}
\end{algorithm}

\section{MNIST-superpixel Settings} \label{app: MNIST-superpixel settings}

\textbf{Datasets} We summarize the statistics of MNIST-superpixel dataset in Table \ref{MNIST-superpixel statistics}. We refer to \citet{MNIST-dataset} for more detailed information if needed.

\begin{table}[h]
\caption{Statistics for MNIST-superpixel dataset.}
\label{MNIST-superpixel statistics}
\begin{center}
\begin{small}
\begin{sc}
\begin{tabular}{c|c|c|c|c}
\toprule
Datasets & Category & Graphs\# & Avg. N\# & Avg. Degree\\
\midrule
MNIST & Superpixel Graphs &  70,000 & 70.57 & 8\\
\bottomrule
\end{tabular}
\end{sc}
\end{small}
\end{center}
\vskip -0.1in
\end{table}

\textbf{Baselines} In semi-supervised MNIST-superpixel experiment, we adopt baselines including: (1) training the model from scratch (\aka \textbf{No Pre-Train}) and that with graph augmentations but without contrastive learning process (\aka \textbf{Aug.}). Following the settings of previous work \cite{GraphCL}, we also take an adjacency information reconstruction method \textbf{GAE} \cite{GAE}, a local \& global mutual information maximization method \textbf{Infograph} \cite{infograph} and a contrastive learning method with a random fashion \textbf{GraphCL} \cite{GraphCL} for comparison.

\section{Transfer Learning Settings} \label{app: transfer learning settings}

\textbf{Datasets} We utilize MoleculeNet \cite{Molecule-Net} as downstream tasks for transfer learning and summarize the statistics which are from \citet{GraphCL} in Table \ref{Transfer learning datasets statistics}. 

\begin{table}[h]
\caption{Statistics for transfer learning MoleculeNet datasets.}
\label{Transfer learning datasets statistics}
\begin{center}
\begin{small}
\begin{sc}
\begin{tabular}{c|c|c|c|c|c}
\toprule
Datasets & Category & Utilization & Graphs\# & Avg. N\# & Avg. Degree\\
\midrule
ZINC-2M & Biochemical Molecules & Pre-training & 2,000,000 & 26.62 & 57.72\\
BBBP & Biochemical Molecules & Finetuning & 2,039 & 24.06 & 51.90\\
Tox21 & Biochemical Molecules & Finetuning & 7,831 & 18.57 & 38.58\\
ToxCast & Biochemical Molecules & Finetuning & 8,576 & 18.78 & 38.52\\
SIDER & Biochemical Molecules & Finetuning & 1,427 & 33.64 & 70.71\\
ClinTox & Biochemical Molecules & Finetuning & 1,477 & 26.15 & 55.76\\
MUV & Biochemical Molecules & Finetuning &93,087 & 24.23 & 52.55\\
HIV & Biochemical Molecules & Finetuning & 41,127 & 25.51 & 54.93\\
BACE & Biochemical Molecules & Finetuning & 1,513 & 34.08 & 73.71\\
\bottomrule
\end{tabular}
\end{sc}
\end{small}
\end{center}
\vskip -0.1in
\end{table}

\textbf{Baselines} In our implementation, we choose various state-of-the-art self-supervised pre-training frameworks as baselines for transferability comparison:

\begin{itemize}[leftmargin=*]
    \item \textbf{Attribute Masking} \cite{GNN-pretrain} Attribute Masking learns the regularities of the node/edge attributes distributed over graph structure to capture inherent domain knowledge.
    \item \textbf{Context Prediction} \cite{GNN-pretrain} Context Prediction predicts subgraphs' surrounding graph structures to pre-train a backbone GNN so that it maps nodes appearing in similar structural contexts to nearby representations. 
    \item \textbf{GraphCL} \cite{GraphCL} GraphCL learns unsupervised representations of graph data via contrastive learning with graph augmentations in a random fashion.
    \item \textbf{GraphLoG} \cite{graph-l-o-g} GraphLoG uses the hierarchical prototypes to capture the global semantic clusters while preserving the local similarities. 
    \item \textbf{AD-GCL} \cite{AD-GCL} AD-GCL optimizes adversarial graph augmentation strategies used in GCL to avoid capturing redundant information. 
\end{itemize}

\section{Unsupervised Learning Settings} \label{app: unsupervised learning settings}

\textbf{Datasets} We summarize the statistics of TU-datasets \cite{TU-dataset} for unsupervised learning in Table \ref{unsupervised learning datasets statistics}. 

\begin{table}[h]
\caption{Statistics for unsupervised learning TU-datasets.}
\label{unsupervised learning datasets statistics}
\begin{center}
\begin{small}
\begin{sc}
\begin{tabular}{c|c|c|c|c}
\toprule
Datasets & Category & Graphs\# & Avg. N\# & Avg. Degree\\
\midrule
NCI1 & Biochemical Molecules &  4,110 & 29.87 & 1.08\\
PROTEINS & Biochemical Molecules &  1,113 & 39.06 & 1.86\\
DD & Biochemical Molecules &  1,178 & 284.32 & 715.66\\
MUTAG & Biochemical Molecules &  188 & 17.93 & 19.79\\
COLLAB & Social Networks &  5,000 & 74.49 & 32.99\\
RDT-B & Social Networks &  2,000 & 429.63 & 1.15\\
RDT-M & Social Networks &  2,000 & 429.63 & 497.75\\
IMDB-B & Social Networks &  1,000 & 19.77 & 96.53\\
\bottomrule
\end{tabular}
\end{sc}
\end{small}
\end{center}
\vskip -0.1in
\end{table}

\textbf{Baselines} To demonstrate the effectiveness of our proposed framework, we adopt multiple strong unsupervised learning baselines, such as Infograph \cite{infograph}, GraphCL \cite{GraphCL} and AD-GCL \cite{AD-GCL} as our baselines. Note that the briefs of aforementioned baselines can be found in Appendix \ref{app: MNIST-superpixel settings} and \ref{app: transfer learning settings}.

\section{Model Structure and Hyperparameters}
To make a fair comparison, we follow the backbone model settings in \citet{GraphCL}. Our model architectures, the mainbody of which includes GCN \cite{GCN} and GIN \cite{GIN}, and corresponding hyperparameters are summarized in Table \ref{Model architectures and hyperparameters}.

\begin{table}[h]
\centering
\caption{Model architectures and hyperparameters}
\label{Model architectures and hyperparameters}
\begin{center}
\begin{small}
\begin{sc}
\begin{tabular}{c|c|c|c|c}
\toprule
Experiments & Transfer Learning & Unsupervised Learning & MNIST-superpixel\\
\midrule
Backbone GNN type & GIN &  GIN & GIN\\
Backbone Neuron\# & [300,300,300,300,300] & [32,32,32] & [110,110,110,110]\\
Rationale Gen. GNN type & GCN &  GIN & GCN\\
Rationale Gen. GNN Neuron\# & [128,64,32] &  [32,32] & [16,16]\\
Rationale Gen. MLP Neuron\# & [32,1] & [32,1]& [16,1]\\
Projector Neuron\# &[300,300] & [32,32] & [110,110]\\
Pooling Layer & global mean pool & global add pool & global add pool\\
\midrule
Learning Rate $lr$ & $\{0.0001,\textbf{0.001},0.01\}$ & $\{0.001,\textbf{0.01},0.1\}$ & $\{0.001,\textbf{0.005},0.01\}$ \\
Sampling Ratio $\rho$ & $\{0.6, 0.7, \textbf{0.8},0.9\}$ & $\{0.7,0.8,\textbf{0.9}\}$ & $\{0.7,\textbf{0.8},0.9\}$\\
Temperature $\tau$ & $\{0.05,\textbf{0.1},0.2,0.3,0.5\}$ & $\{0.1,\textbf{0.2},0.3\}$ & $\{0.1, 0.2, 0.3, \textbf{0.5},0.7\}$\\
$\lambda$ in Equation \ref{equ:final_loss_function} & $\{0.05,\textbf{0.1}, 0.3, 0.5, 1.0\}$ & $\{\textbf{0.1}, 0.5, 1.0\}$ & $\{0.1, \textbf{0.2},0.5\}$\\
Training Epochs & $\{60,\textbf{80},100\}$ & $\{10,\textbf{20},40\}$ & $\{60,\textbf{100},150\}$\\
\bottomrule
\end{tabular}
\end{sc}
\end{small}
\end{center}
\vskip -0.1in
\end{table}

Note that $\{\cdot\}$ indicates the grid research ranges we tune these hyperparameters in and \textbf{bold} numbers are our final settings. 

\section{Complexity Analysis}
We make a comparison on complexity between RGCL and GraphCL in this section. In terms of space complexity, besides the backbone encoder, although RGCL introduces another rationale generator to capture rationales in anchor graphs, space complexity will not be significantly increased since the rationale generator is a much smaller network compared with the backbone encoder and its scale can be adjusted if necessary.

Then we move on to time complexity analysis of pre-training process. Suppose the average numbers of nodes and edges per graph in pre-training datasets are $|\mathcal{V}|$ and $|\mathcal{E}|$, respectively. Let $B$ denote the batch size, $\rho$ denote the sampling ratio, $L_B$ denote the number of GNN layers in backbone encoder, $L_R$ denote that in rationale generator and $d$ denote the latent space dimension where contrastive loss is calculated. The time complexity comparison of processing a minibatch of training data is presented in Table \ref{Time complexity comparison}. 

\begin{table}[h]
\centering
\caption{The comparison of time complexity between RGCL and GraphCL}
\label{Time complexity comparison}
\begin{center}
\begin{small}
\begin{sc}
\begin{tabular}{c|c|c|c|c}
\toprule
Frameworks & GraphCL & RGCL\\
\hline
\hline
Data Augmentation & $O(2B\rho |\mathcal{V}| \log |\mathcal{V}|)$ & $O(3B\rho |\mathcal{V}| \log |\mathcal{V}|)$ \\
\hline
Rationale Generator Propagation & --- & $O((|\mathcal{E}|^2+|\mathcal{V}|)L_RB)$ \\
\hline
Backbone Encoder Propagation & $O((2|\mathcal{E}|^2+|\mathcal{V}|)L_BB)$ & $O((3|\mathcal{E}|^2+|\mathcal{V}|)L_BB)$ \\
\hline
Contrastive Loss & $O(B^2d)$ & $O(2B^2d)$ \\
\bottomrule
\end{tabular}
\end{sc}
\end{small}
\end{center}
\vskip -0.1in
\end{table}

As illustrated in Figure \ref{fig:framework}, RGCL has a three-tower structure while GraphCL has a two-tower one. The extra one stems from the calculation of the complements of rationales to regulate the information outside the rationale. And in realistic implementation on our platform (GeForce RTX 2080 Ti and Intel(R) Core(TM) i9-9900X), it takes about 15 hours to finish GraphCL pre-training while 20 hours are needed for RGCL, which roughly matches our theoretical analysis above.









\end{document}